\def\year{2020}\relax
\documentclass[letterpaper]{article} 
\usepackage{aaai20}  
\usepackage{times}  
\usepackage{helvet} 
\usepackage{courier}  
\usepackage[hyphens]{url}  
\usepackage{graphicx} 
\usepackage{graphicx}  
\usepackage{epsfig}
\usepackage{amsmath}
\usepackage{amssymb}
\usepackage{latexsym}
\usepackage{color}

\usepackage{amsthm} 
\usepackage{graphics} 
\usepackage{algorithmic} 
\usepackage{subfig} 
\usepackage{url} 
\usepackage{kotex}
\usepackage{bm}
\usepackage{adjustbox}
\usepackage{multirow}

\usepackage{array}
\usepackage{tabu}

\usepackage{amsthm} 
\usepackage{algorithmic} 
\usepackage{subfig} 
\usepackage{kotex}
\usepackage{bm}
\usepackage{adjustbox}
\usepackage{multirow}

\usepackage{array}
\usepackage{tabu}
\usepackage{ifsym}

\title{Tell Me What They're Holding: Weakly-Supervised Object Detection \\
with Transferable Knowledge from Human-object Interaction
}
\author{Daesik Kim$^{1,2,*}$ \quad Gyujeong Lee$^{1,*}$ \quad Jisoo Jeong$^{1,*}$ \quad Nojun Kwak$^{1,\dagger}$ \\
$^{1}$Seoul National University \quad $^{2}$V.DO Inc\\
\tt\small{\{daesik.kim|regulation.lee|soo3553|nojunk\}@snu.ac.kr }
}

\newcommand{\gl}[1]{\textcolor{black}{#1}}

\newcommand{\nj}[1]{\textcolor{black}{#1}}

\begin{document}

\maketitle

\begin{abstract}
In this work, we introduce a novel weakly supervised object detection (WSOD) paradigm to detect objects belonging to rare classes that have not many examples using transferable knowledge from human-object interactions (HOI).
While WSOD shows lower performance than full supervision, we mainly focus on HOI as the main context which can strongly supervise complex semantics in images.
Therefore, we propose a novel module called RRPN (relational region proposal network) which outputs an object-localizing attention map only with human poses and action verbs.
In the source domain, we fully train an object detector and the RRPN with full supervision of HOI.
With transferred knowledge about localization map from the trained RRPN, a new object detector can learn unseen objects with weak verbal supervision of HOI without bounding box annotations in the target domain.
Because the RRPN is designed as an add-on type, we can apply it not only to the object detection but also to other domains such as semantic segmentation. 
The experimental results on HICO-DET dataset show the possibility that the proposed method can be a cheap alternative for the current supervised object detection paradigm. 
Moreover, qualitative results demonstrate that our model can properly localize unseen objects on HICO-DET and V-COCO datasets.

\end{abstract}

\section{Introduction}

{\let\thefootnote\relax\footnotetext{{
*All of authors equally contributed to this work.}}\footnotetext{{
$\dagger$Corresponding author.}}}

In a decade, object detection has become one of the most successful 
fields in computer vision with various applications \cite{Ren15,dai2016r,redmon2016you,liu2016ssd}.
Most of the successful models have emerged after the release of large scale datasets (e.g. PASCAL VOC, MS-COCO \cite{everingham2010pascal,lin2014microsoft}) with bounding box annotations.
Given input images, conventional object detection models can localize boxes \nj{with the corresponding class scores.} 
Thus, they normally require manually annotated bounding boxes \nj{containing} accurate coordinate values and object labels for training.

\begin{figure}[t]
\centering
\includegraphics[width=\linewidth]{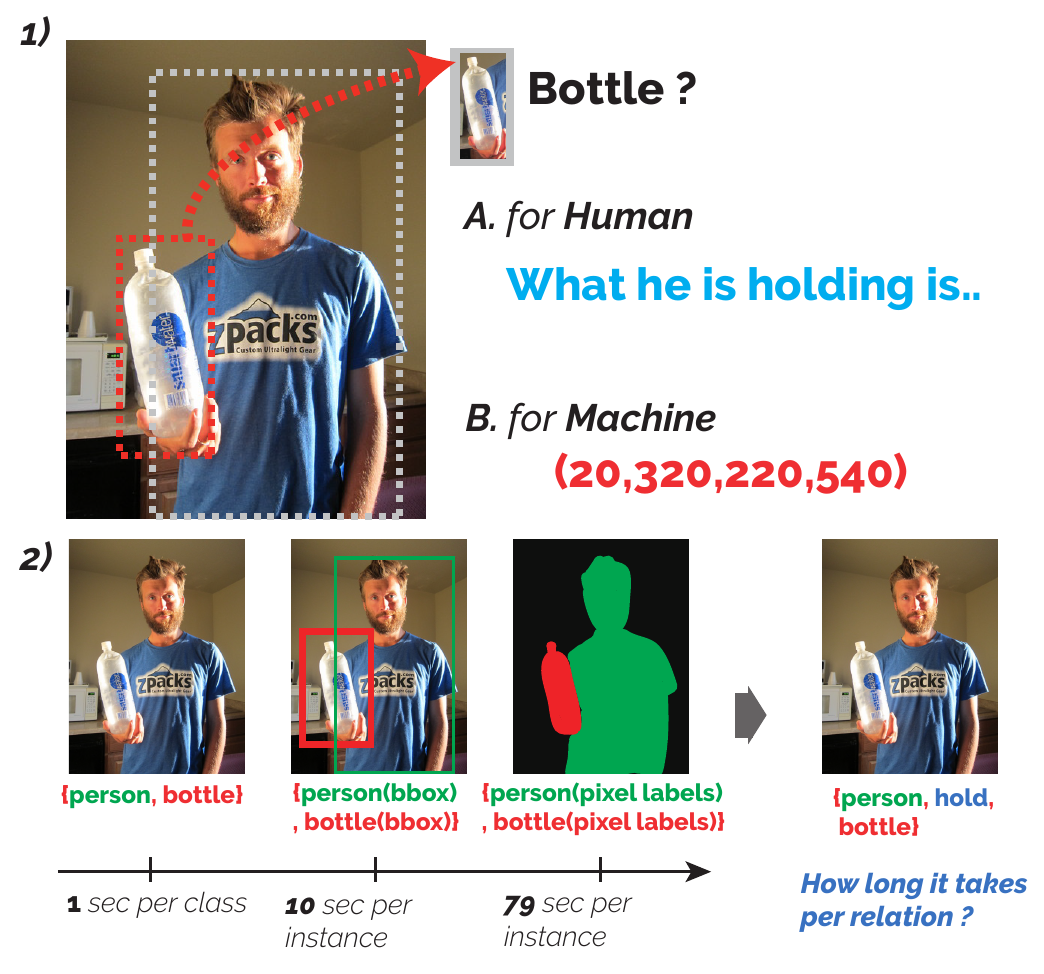}
\caption{1) Two different types of description of an object. A is human's way of identifying an object while B is for machines.
2) Manually annotating time for three tasks. Bearman et al. \cite{bearman2016s} estimated annotation times for image level, bounding box and pixel level. At rightmost, annotating time of relation sentence can be similar to that of image-level since only action verb ``hold" is added. }
\label{fig:Intro}
\vspace{-2mm}
\end{figure}

However, annotating bounding boxes is time-consuming and labor-intensive.
It can also be difficult to expand the volume of a dataset by adding more object classes or adding more images.
Therefore, researches to reduce those costs in various ways have drawn attentions these days.

Weakly supervised object detection (WSOD) has been proposed to tackle \nj{the} aforementioned problems \cite{zhu2017soft,shi2017weakly,jie2017deep}.
\nj{It} is to detect objects within images by weak supervision such as image-level labels.
\nj{At the cost of lowered annotation cost, WSOD performs worse than}  full supervision.

To overcome a limitation of weak supervision, some approaches \cite{shi2017weakly} rely on another type of full supervision with transfer learning.
Transferred knowledge from a source domain could support weak supervision in a target domain.
However, annotating other types of \nj{labels} 
such as segmentation mask 
is also expensive.

Our main intuition is that supervision of machines is totally different from that of humans.
For example, Fig. \ref{fig:Intro}(1) shows different ways of identifying objects between humans and machines.
While we should provide accurate coordinate values of object boxes for machines, \nj{humans} usually \nj{recognize} new objects from contexts.
Contexts also can reinforce supervision without much additional efforts.

Especially, how objects are related to human actions can be practical and advantageous since information about a human can be a proper evidence for recognizing contexts in an image. Moreover, humans can easily express contexts with sentences as shown in Fig. \ref{fig:Intro}(1), so that linguistic labels can be a key to reduce annotation cost for humans \nj{as} shown in Fig. \ref{fig:Intro}(2).
\nj{Compared} to other annotating costs \cite{bearman2016s}, \nj{the} cost of annotating a relation sentence such as ``person, hold, bottle" can be almost similar to that of image-level annotation. 
Thus, we propose a novel paradigm to learn unseen objects based on human-object interaction (HOI). 

Our key idea is to exploit transferable knowledge from HOI contexts annotated \nj{by} language as is in \cite{chao2018learning}.
\nj{Specifically,} we propose a novel module that predicts object locations from HOI.
Since the actual coordinate values can not be specified, we use an attention map as localization results to connect it with a bounding box.
Moreover, in order to train \nj{a} full object detector (e.g. Faster-RCNN \cite{Ren15}) in an end-to-end fashion, we design a new module as an add-on type.

The objective of this paper is to make our model learn additional rare classes with weak verbal \nj{supervision} annotated easily by human.
During the first stage, strong \nj{supervision} on non-rare classes teach our model to localize a proper location with a human pose and an action verb.
In the next stage, only weak \nj{supervision} with transferred knowledge keep\nj{s} training \nj{an} object detector for unseen rare classes.

Our main contributions can be summarized as follows:
\begin{itemize}
\vspace{-1mm}
\item We define a new weakly-supervised object detection scheme which mainly relies on interactions between a human and objects without box annotations.

\vspace{-1mm}
\item We propose a novel module called RRPN (Relational Region Proposal Network) to localize boxes by using the location of a human and verb embeddings. 

\vspace{-1mm}
\item The proposed RRPN is designed as a universal add-on type that can be easily adapted into existing models such as Faster-RCNN.
\end{itemize}

Our experiments validate that our model outperforms baselines on the HICO-DET \cite{chao2018learning} dataset
and can effectively transfer the knowledge to \nj{other} dataset such as V-COCO \cite{gupta2015visual}.

\section{Related works}
\noindent \textbf{Weakly Supervised Object Localization and Detection}
Most of the weakly supervised object localization and detection methods have been proposed based on an image-level supervision. 
With cheaper but weaker annotations, studies \cite{bilen2016weakly,diba2017weakly,kantorov2016contextlocnet,oquab2015object,tang2017multiple,jie2017deep} mainly tried to enhance performance by multiple instance learning (MIL).
In MIL, a bag is defined as a collection of regions in an image. It is labeled as positive if at least one object is positive and labeled as negative if all the objects are negative. 

\citeauthor{tang2018pcl} \citeyear{tang2018pcl} have proposed proposal cluster learning algorithm to learn refined instance classifier. \citeauthor{yang2019activity} \citeyear{yang2019activity} have proposed activity-driven WSOD, which also exploits action classes as contextual information to localize objects without box annotations. However, those works has generated box proposals using Selective Search \cite{uijlings2013selective}, which is a rule-based algorithm. Since the box proposal method cannot be trained, there is a fundamental limitation which proper proposals hardly exist in novel data. \citeauthor{uijlings2018revisiting} \citeyear{uijlings2018revisiting} have addressed a new WSOD framework that revisits knowledge transfer for training object detectors on target classes. Since this work has optimized a box proposal network for target classes by MIL, box generators can be insufficiently trained with rare classes due to a lack of contextual information. Our method resolves the aforementioned issues with a novel box proposal module that can transfer knowledge using an HOI dataset.

\noindent \textbf{Human-object interaction}
Visual recognition of HOI is crucial for comprehending a scene in an image.  
Early work studied mutual context of human pose and objects \cite{yao2010modeling} and Bayesian model \cite{gupta2007objects,gupta2009observing} with handcraft features.
Recently, with success of \nj{deep learning}, \citeauthor{chao2015hico} \citeyear{chao2015hico} 
introduced a new large-scale benchmark, ``Humans Interacting with Common Objects'' (HICO), for HOI recognition, which was expanded for detection problems in HICO-DET \cite{chao2018learning}.
In order to solve HICO-DET datasets, various approaches have been proposed. In \cite{chao2018learning}, combined features from human proposals and object regions were used to solve HOI detection. 
\citeauthor{gkioxari2018detecting}
\citeyear{gkioxari2018detecting} proposed a human and object detector-based approach estimating a density map based on Faster-RCNN architecture. A recent approach \cite{qi2018learning} generates the HOI graph and propagates message between nodes to infer relationships in a parsing graph. In this paper, rather than directly solving HOI problems, we exploit contextual information in HICO-DET to construct a weakly-supervised object detector.

\section{Algorithm Overview}
\label{sec:method}

\begin{figure}[t]
\centering
\includegraphics[width=\linewidth]{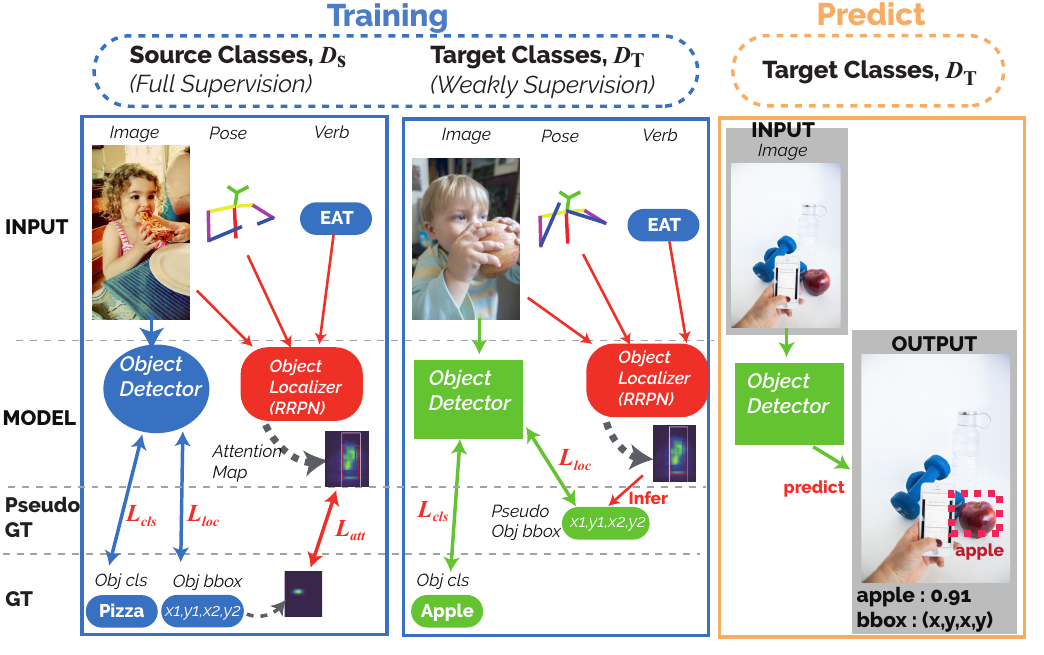}
\caption{\textbf{Overview of our algorithm.} 
1) During the training phase for source classes, RRPN is also trained to predict attention map from human-object interaction. 
2) In the target-class training phase, an object detector is trained using the ground truth class label, and the box label provided by the trained RRPN. 
In other words, our problem focuses solely on solving the weakly supervised object detection problem on the target classes.
3) As a result, the trained object detector for target classes can infer box coordinates and object classes with only an image input.}
\label{concept_img}
\vspace{-2mm}
\end{figure}

An overview of our algorithm is illustrated in Fig. \ref{concept_img}.
Let $D =  \{ (I_{i}, y_{i})  \}_{i=1}^{N}$ is the data set,
where $y_{i}$ is the label of the image $I_{i}$ and $N$ is the number of images.
The image label $y_{i}$ is organized as a tuple as shown below:
\begin{align}\label{eq:tuple} 
    y_{i} =  \{  ( H_{verb}^{j}, O_{bbox}^{j}, O_{cls}^{j} )  \}_{j=1}^{M}
\end{align}
where, $O_{bbox}^{j}, O_{cls}^{j}$ are the bounding box and the class of an object in the image, 
$H_{verb}^{j}$ is the action verb corresponding to the object,
and $M$ is the number of tuples in the image $I_{i}$.
To evaluate the proposed method, we divided $D$ into two sub-categories based on the number of objects in a class: 
non-rare (source classes) $D_{S}$ and rare (target classes) $D_{T}$.
Note that, there is no object class duplicates but all action verbs are overlapped between two subcategory datasets.

For $D_{S}$, we normally train the first object detector (blue circle in Fig. \ref{concept_img}) with full supervision using ($O_{bbox}^{j}, O_{cls}^{j}$). Along with training of the object detector, we also train an object localizer (red circle) called RRPN with newly defined inputs. Since the RRPN should learn how to localize an object only with the information on a human and an action verb, we use the image $I_{i}$, the verb $H_{verb}^{j}$, and the pose of the human $H_{pose}^{j}$ as inputs. The $H_{verb}^{j}$ simply comes from $y_{i}$ but the human pose $H_{pose}^{j}$ is extracted from an image $I_{i}$ with an existing human pose estimation method.
As a results, the RRPN predicts an attention map $\tilde{A}_{i}^j$ of an object location in the $i$-th image from human's action and appearance.
We optimize losses regarding the object class and the location using $O_{bbox}^{j}$ and $O_{cls}^{j}$ for the object detector, but create a Gaussian map of $O_{bbox}^{j}$ and use it as a ground truth in the training of the attention map of the RRPN.
In this phase, since 
\nj{the ground truth bounding box location is available,}
the RRPN can learn common knowledge between objects and human actions.


For $D_{T}$, \nj{we assume that} only object class information $O_{cls}^{j}$ and the action verb $H_{verb}^j$ are available but \nj{the bounding box information is not}. 
To fill the absence of $O_{bbox}^{j}$, we exploit learned knowledge inferred by the RRPN with the same kinds of inputs as the training phase for the source classes. 
Since the output of RRPN is an attention map, we extract a coordinate by thresholding it and generate pseudo bounding box $\hat{O}_{bbox}^{j}$. 
Then, we normally train the second object detector (green rectangle in Fig. \ref{concept_img}) for $D_{T}$. 
Since we already have used all action verbs to train the RRPN in the previous phase and transfer the same parameters in the training phase for the target classes, it can infer an object location with a human pose and an action verb.
In Fig. \ref{concept_img}, after the RRPN already learned to localize unseen object ``\textit{Apple}" with verb ``\textit{EAT}" and grabbing pose in the training phase of $D_{S}$, it can infer \nj{a} proper location as a pseudo ground truth $\hat{O}_{bbox}^{j}$. 
In conclusion, we use weak supervision by human actions to train a full object detector. 

Eventually, the trained object detector in the second phase can predict objects in $D_{T}$ only with an input image $I_{i}$ as shown in Fig. \ref{concept_img}. Although we have not shown the real location of ``\textit{Apple}", it is possible to predict the class score and the coordinate of an ``\textit{Apple}" object. 

\begin{figure*}[ht]
\centering
\includegraphics[width=0.85\linewidth]{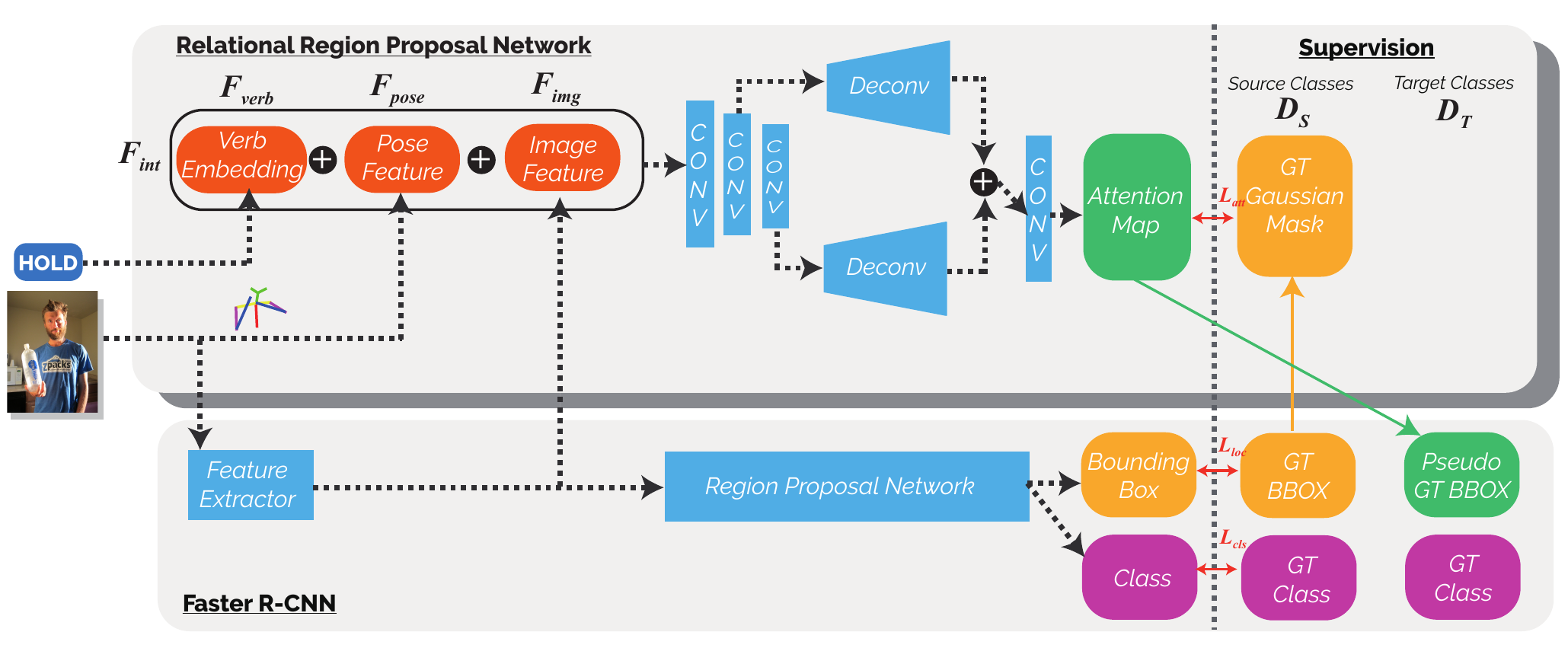}
\caption{Overall network architecture of the proposed algorithm. Relational Region Proposal Network (RRPN) at the top is mounted on a basic Faster-RCNN model at the bottom. In RRPN, a combined feature $F_{int}$ from verb, pose and image produces an attention map through a network which has four blocks. With source classes, the RRPN is trained with a Gaussian mask from the ground truth bounding box. However, with target classes, the RRPN generates a pseudo ground truth bounding box so that Faster-RCNN can be optimized.}
\label{architecture}
\end{figure*}

In this scheme, we can additionally train new object detector for unseen rare classes without bounding box annotations.
Moreover, since we already trained the RRPN with strong supervisions, we need smaller amount of data in target classes compared to other WSOD algorithms.
Our experiments validate that \nj{our target domain which contains extremely rare object classes is trained successfully by our method.}




\section{Architecture}

Fig.~\ref{architecture} depicts the overall architecture of the proposed algorithm.
The proposed algorithm consists of two modules,
including the RRPN and the object detector.
More precisely, it means that RRPN can be combined with the conventional architecture such as the Faster-RCNN.
RRPN is a multi-stage encoder-decoder network, which is responsible for predicting an object-location-centric attention map $A_{i}^j$
from a multi-domain integrated feature map. 
The other module, object detector, is a conventional object detector
which is trained for a given input image using the ground truth label $O_{cls}$ and the bounding box $O_{bbox}$ .

In order to exploit the knowledge of interaction, we train $D_{T}$ after the training of $D_{S}$ is done.
Since, however, we do not account for the continual learning,
$D_{S}$ and $D_{T}$ do not share parameters for the object detector.
While the object detector is trained for $D_{S}$ with supervision, at the same time,
RRPN is also trained to learn the knowledge from interactions between a human and an object through  action verbs. 
Then, the object detector is trained for $D_{T}$ without object bounding boxes, i.e. in a weakly supervised way, 
using the transferred knowledge from $D_{S}$.



\subsection{Training on the Source classes $D_S$}
\label{sec:Ds}
$D_{S}$ are object classes on which data can be easily acquired.
Training on $D_{S}$ is a standard supervised object detection procedure by using ground truth class and box labels for all the objects.
The main purpose of training on $D_{S}$ is to predict an object location from a human-object interaction.
Therefore, RRPN is also trained at the same time as the training of the object detector. 
The detailed training procedure for Faster-RCNN is applied in the same way as the original paper.
The training procedure of RRPN is as follows.

\subsubsection{Relational Region Proposal Network (RRPN)}
RRPN is designed to be universally applicable to various task's models, including other object detectors, 
in an add-on manner, and can share the backbone network with other model for image features to improve memory efficiency.

As mentioned above, RRPN predicts attention map $\tilde{A}_{i}^j$ for a given image $I_{i}$ using a multi-domain feature map $F_{int}^{i,j}\in\mathbb{R}^{C\times H\times W}$ as an input,
where $C, W, H$ are the depth, width and height of the feature map. 
$F_{int}$ 
is obtained by 
\begin{align}\label{eq:int_featmap} 
   F_{int}^{i} = \{F_{int}^{i,j}\}_{j=1}^ M = \{ ( F_{img}^{j} \oplus F_{pose}^{j} \oplus F_{verb}^{j}  ) \}_{j=1}^{M},
\end{align}
where, $F_{img}$, $F_{pose}$, $F_{word}$ are the image feature, pose feature, and verb feature
obtained by their corresponding models $f_{img}(I_{i}, \theta_{img})$, $f_{pose}(I_{i}, \theta_{pose})$,  
$f_{word}(H_{act}^{j}, \theta_{verb})$, and $\oplus$ is the matrix concatenation.
Here, $\theta_x$ is the corresponding model parameters.
The convolution operation for $F_{int}^{i}$ computes the object existence probability
for a combination of $F_{img}^{j}$, $F_{pose}^{j}$, and $F_{verb}^{j}$ at a specific location on $I_{i}$.

As in (\ref{eq:int_featmap}), we used three feature maps to utilize contexts from various domains in a given dataset, 
and each feature map has its own contribution.
The pose and word feature are responsible for the visual context of the human's location and action,
and the distinguishable linguistic context for the human's action, respectively. 
The image feature is responsible for representing the whole scene as well as the object of interest.
The details for each feature maps are as follows:

\noindent \textbf{Pose feature} We use the well-known human pose estimation model, OPENPOSE \cite{cao17}, to extract pose features.
OPENPOSE predicts the location of human body joints using 
image or video as an input.
The output consists of channels corresponding to each joint and a channel representing background information.
In this paper, we used a pose estimation model with 19 channels including 18 joints and 1 background. 
In order to feed distinct information of human pose to the RRPN, 
we exploit the 18 channels except for the background channel as the pose feature.


\noindent \textbf{Verb feature} The widely used GloVe-twitter-27B-25d model \cite{Pennington14} is applied as the word embedding model for the verb.
Since a word 
is embedded into a vector, 
one needs to convert it into a tensor form for integration with other features.
While $F_{img}^{j}$ and $F_{pose}^{j}$ may have different spatial-wise activations depending on $I_i$, 
$F_{verb}^{j}$ must have the same value regardless of positions.
In designing $F_{verb}^{j}$, we also take this consideration into account.
In order to match the spatial dimension with others, the verb feature is copied to every spatial position. 
So that dimension of $F_{verb}^{j}$ is converted from $\mathbb{R}^{25}$ to $\mathbb{R}^{25\times H\times W}$.
By stacking a depth-wise word vector at all spatial positions, 
we can conduct a convolution operation using the same verbal information 
at all position of $F_{img}^{j}$.
Note that, among the HICO-DET datasets, 
tuples with `No interaction' verb labels were excluded from training and validation phases
for accurate evaluation of the proposed algorithm.

\noindent \textbf{Image feature} The proposed algorithm makes use of the representative two-stage object detection model, Faster-RCNN. 
It consists of a feature extractor, a back-bone network, and a region proposal network (RPN). 
The output feature map of the back-bone network of the Faster-RCNN is used as the image feature for the RRPN. 
When training on the target classes, the parameters of the backbone network are reused, but the parameters of RPN are reset.



Multi-domain feature map $F_{int}^{i,j}$ is then fed into the network to predict attention map $A_{i}^j$.
In order to robustly detect objects in various sizes, 
we designed the network architecture which has four blocks as in \cite{Wang18}: an Encoder block $f_{en}$, two decoder blocks $f_{de1}$, $f_{de2}$ and an attention block $f_{att}$.
$f_{en}$ takes $F_{int}^{i,j}$ as an input and outputs two feature maps with different spatial dimensions.
Then, each output feature map feeds into $f_{de1}$ and $f_{de2}$, respectively.
The output feature maps of $f_{de1}$ and $f_{de2}$ having the same spatial dimension are concatenated and inputted to the attention block resulting in an attention map $\tilde{A}_{i}^j$
as 
\begin{align}\label{eq:decov} 
   \tilde{A}_{i}^j = f_{att} [f_{de1} \{f_{en}^{1}(F_{int}^{i,j}) \} \oplus f_{de2} \{f_{en}^{2}(F_{int}^{i,j}) \} ].
\end{align}

The output of RRPN is an attention map which emphasizes the location where the object \nj{is} likely to be located.
%
To train attention maps, we create a Gaussian map $A_{i}^j$, as a ground truth attention map, using $O_{bbox}^{j}$.
RRPN is trained using $A_{i}^j$ as the label. 
We use pixel-wise binary cross entropy loss (BCE) $L_{att} $ between $(A_{i}^j, \tilde{A_{i}^j})$. 

The total loss for training on the source classes including RRPN and object detector is shown below:
\begin{align} 
\begin{split}
    L_{total} = L_{det}+\lambda L_{att}, \quad
    L_{det} = L_{cls}+L_{loc}
\end{split}
\label{eq:total_loss}
\end{align}
where, $\lambda$ is a hyper-parameter balancing between the two losses and 
$L_{det}$ is the loss for the Faster-RCNN.
%
In the object detector point of view, the proposed algorithm on $D_{S}$ 
is trained in the same way as the conventional supervised objected detection algorithms.

\subsection{Training On the Target classes $D_T$}
\label{sec:Dt}
The object detector for $D_T$ should be trained without $O_{bbox}$.
Therefore, we define this problem as a weakly supervised object detection (WSOD) problem.
We use $\tilde{O}_{bbox}$ as an alternative to the missing $O_{bbox}$ 
utilizing RRPN learned in the source classes training phase.
It is expected that the trained RRPN can predict locations of unseen objects 
i.e. $D_T$,  
since it is trained to predict the object location using a human pose, an action (verb) and an image feature.
The training process on $D_T$ using the trained RRPN is as follows:

The $F_{pose}$, $F_{verb}$, and $F_{img}$ are fed into the trained RRPN.
We apply a threshold to obtain a pseudo bounding box from the output attention map as 
\begin{align}\label{eq:threshold} 
\tilde{A}_{i}^{j} = \begin{cases}
  \: \: 1, &\textup{if}, \: \: \tilde{A}_{i}^{j} > \delta \\
  \: \: 0, &\textup{otherwise}
\end{cases}
\end{align}
where, $\delta$ is a pre-defined threshold.
The largest bounding box containing a valid value in $\tilde{A}_{i}^{j}$ is called $\tilde{O}_{bbox}^{j}$.
The pseudo ground truth bounding boxes $\{\tilde{O}_{bbox}^{j}\}_{j=1}^M$ obtained from 
the attention maps $\{\tilde{A}_{i}^{j}\}_{j=1}^M$ of all tuples in the image $I_i$
are collected together and used as bounding box labels $\tilde{O}_{bbox}$ for training an object detector. In this step, a different type of object detector from the one trained in $D_{S}$ training phase can be used for training.
The object detector \nj{is trained} to minimize detection loss using $O_{cls}$ and $\tilde{O}_{bbox}$.


\section{Experiment}

In this section, we evaluate the performance of the proposed WSOD algorithm.
To the best of our knowledge, no previous studies have been conducted on the relationship between object detection and HOI. 
Nevertheless, we conduct\nj{ed} the performance comparison with prior works on HICO-DET.

\subsection{Dataset and Pre-processing}

HICO-DET dataset consists of 47,776 images (38,118 training and 9,658 testing) classified into 117 actions (verb) and 80 object classes, 
and the object classes are the same as MS-COCO dataset. 
The ground truth labels consist of a tuple of $(H_{act}$, $O_{bbox}$, $O_{cls})$ as in (\ref{eq:tuple}). 
Note that, the RRPN is trained based on tuples, so images containing multiple tuples are fed multiple times.
The total number of tuples is 151,276 (117,871 training and 33,405 testing), and we use 131,560 tuples (102,450 training and 29,110 testing) excluding the tuples corresponding to the action label `no-interaction'.

In order to construct the problem environment, 
the whole dataset is divided into source and target datasets according to the frequency of the object class. 
Our basic experiment is set up with 116 verbs excluding `no interaction'. \nj{The number of object classes are 70 for $D_{S}$ and 10 for $D_{T}$.} 
In order \nj{to more clearly show} the effectiveness of RRPN, \nj{in the experiment for qualitative result,} we use 5 verbs and 10 $D_{S}$ and 70 $D_{T}$. \nj{O}ther hyperparameters remain the same as the basic set up.


We also verified the proposed algorithm on V-COCO dataset for qualitative analysis.
The purpose of evaluation on the V-COCO dataset is to show that the knowledge can be transferred from one dataset to other. 
Details on both datasets are described in the supplementary material.



\subsection{Metrics}
We use mean Average Precision (mAP) and Recall as evaluation metrics.
Because RRPN produces one bounding box for one tuple (action), 
Recall is used to measure how accurate the location of an object corresponding to an action is.
In other words, Recall evaluates the objectness of $\tilde{A}$ predicted by RRPN, and is calculated as the ratio of tuples for which IoU $> 0.5$.
On the other hand, the object detector detects all the objects in an image at once. Therefore, we use the mAP in measuring the performance of Faster-RCNN which are the standard metrics for object detectors.

$D_S$ and $D_T$ in Recall are the performance of the RRPN's agent after the training on $D_S$. When training on $D_T$, 
RRPN is fixed and not trained. 
Note that Recall is measured on test set for $D_{S}$ and on both training and test set for $D_{T}$.

\subsection{Comparison with prior works}
We conduct experiments to compare with prior works on HICO-DET as shown in Table~\ref{exp:wsod}.
First two columns represent overall results of original algorithms in AD~\cite{yang2019activity} and PCL~\cite{tang2018pcl}.
However, both results are only able to show performance of all object classes with an entire dataset.
Since our method is designed for transfer learning, we experiment to validate PCL on each of source and target domains. 
As a result, our best model with image, pose and verb has 17.19\% which is 4 times better than \nj{the} result of PCL on $D_{T}$.
Moreover, our model only with the image feature outperforms PCL on $D_{T}$.
Although a direct apple-to-apple comparison is difficult, we can see that our method is far better than the compared methods.

\begin{table}[t]
\caption[Comparison of quantitative results with prior works]{Comparison of the mAP with other WSOD algorithms on HICO-DET. (PCL* is tested by ourselves, {$\S$} is trained on the entire dataset and {$\dagger$} is trained on $D_{S}$ and $D_{T}$ separately. I : $F_{img}$, P : $F_{pose}$, V: $F_{verb}$, W : Weakly supervised object detection, $\lambda$ = 10, $\delta$ = 0.1)}
\centering
\begin{adjustbox}{max width=\linewidth}
\vspace{0.5cm}
\renewcommand{\tabcolsep}{2pt}
\begin{tabular}{| l || c | c | c | c || c | c |}
\hline
Methods & AD & PCL  & \multicolumn{2}{|c||}{PCL*}
  & Ours (I) &  Ours (I+P+V) \\
\hline
$D_{S}(W)$ & - & - & $4.80^{\S}$ & $5.01^{\dagger}$  & - & - \\
$D_{T}(W)$ & - & - & $0.01^{\S}$      & $4.75^{\dagger}$             & 9.57 & 17.19 \\
\hline
Total & 5.39 & 3.62 & $4.42^{\S}$ & -               & - & - \\

\hline
\end{tabular}

\end{adjustbox}
\label{exp:wsod}

\end{table}

\begin{table}[ht]
\caption{Performance comparison of 
different feature combination. (Notations, $\lambda$, and $\delta$ are the same as Table~\ref{exp:wsod})}

\vspace{-1mm}
\centering
\begin{adjustbox}{max width=\textwidth}
\begin{small}
\renewcommand{\tabcolsep}{4pt}

\begin{tabular}{| c | c | c || c | c || c | c | c |}
\hline
I & P & V & \multicolumn{2}{c||}{\footnotesize{Recall@.5} \scriptsize{(RRPN)}}       & \multicolumn{3}{c|}{mAP@.5 \footnotesize{(Faster-RCNN)}} \\
\cline{4-8}                    & &     & $D_{S}$ ($\%$) & $D_{T}$ ($\%$) & $D_{S}$         & $D_{T}$(W)  & $D_{T}$(S)\\
\hline
\hline
\checkmark & \checkmark & \checkmark    & \textbf{47.69}    & \textbf{28.64}        & \textbf{30.34}  & \textbf{17.19} & 29.37  \\
\hline
\checkmark &  &                         & 42.00    & 22.75        & 23.57  & 9.57 & 22.07   \\
\hline
\checkmark & \checkmark &               & 41.42    & 22.13        & 24.17  & 10.07 & 25.15 \\
\checkmark &  & \checkmark              & 46.34    & 23.84        & 29.97  & 16.34 & \textbf{30.28}   \\
\hline
\end{tabular}
\end{small}
\end{adjustbox}
\label{exp:hico_ablation}
\end{table}

\subsection{Comparison with different feature combination}

We experiment to verify the performance of different feature combinations. We train and test the RRPN using the same types of feature for both $D_{S}$ and $D_{T}$ in each experiment. Table~\ref{exp:hico_ablation} shows the performances of RRPN and Faster-RCNN as Recall and mAP, respectively, using different combinations of features. $D_{T}(W)$ in mAP is the results of our WSOD, and $D_{S}$ and $D_{T}(S)$ are the results of full supervision. 

Our full model combining all three features in the top of Table~\ref{exp:hico_ablation} shows the highest performance in both Recall and mAP among all combinations. 
The mAP of $D_{T}(W)$ has 17.19\%, which is 7.62\% better than image-only model and the Recall of the $D_{T}$ is 28.64\% which is about 5\% higher than other combinations.
Moreover, the mAP score of our full model is only 4.88\% lower than image-only fully supervised model in $D_{T}(S)$. 
Compared to models of full supervision $D_{T}(S)$, we believe that the mAP score of our full model can meaningfully show that it can be trained despite weak supervision of rare classes. 


In the middle, using $F_{img}$ alone, Recall for $D_{T}$ has 22.75\% 
and mAP ($D_{T}(W)$) are much lower than mAP ($D_{T}(S)$). 
It means that RRPN could not be trained solely by $F_{img}$. 
The two results in the bottom are the performance for combined features. 
When $F_{pose}$ is combined with $F_{img}$, Recall degrades and mAP increases slightly. 
It can show that $F_{pose}$ that is extracted from an image is redundant unless it interacts with a verb. 
Combining $F_{verb}$ with $F_{img}$, however, 
Recall and mAP significantly increase and mAP($D_{T}(S)$)
also increases. 
It is interesting that it might be more effective for not only RRPN but also Faster-RCNN when using combination of features from other domain.




\begin{table}[h]
\caption{Comparison of quantitative result of $\lambda$ and $\delta$
(Notations are the same as Table~\ref{exp:wsod})}
\vspace{-1mm}
\centering
\begin{adjustbox}{max width=\textwidth}
\renewcommand{\tabcolsep}{5pt}
\begin{small}
\begin{tabular}{|c | c || c | c || c | c | c |}
\hline
\multicolumn{2}{|c||}{parameter}  &\multicolumn{2}{c||}{\footnotesize{Recall@.5} \scriptsize{(RRPN)}}       & \multicolumn{3}{c|}{mAP@.5 \footnotesize{(Faster-RCNN)}} \\
\hline
$\lambda$ & $\delta$    & $D_{S}$ ($\%$) & $D_{T}$ ($\%$)  & $D_{S}$         & $D_{T}$(W)  & $D_{T}$(S) \\
\hline
\hline
0 & 0.1        & 11.64    & 11.19        & \textbf{31.06}  & 1.61 & 25.57 \\
\hline
 1 & 0.1                               & 41.51    & 24.46        & 23.87  & 9.27 & 25.43   \\
 5 & 0.1                              & 46.37    & 23.37        & 30.10  & 14.38 & 26.11  \\
 10 & 0.1                             & \textbf{47.69}   & \textbf{28.64}       & 30.34  & \textbf{17.19} & \textbf{29.37} \\
 15 & 0.1                             & 46.65    & 23.07        & 23.32  & 15.85 & 25.45   \\
 20 & 0.1                             & 43.17    & 26.00        & 22.68  & 15.75 & 22.92  \\
\hline
\hline
 10 & 0.05     & \textbf{48.22}    & \textbf{29.41 }       & 30.27  & 9.41 & 25.04   \\
 10 & 0.10                          & 47.69    & 28.64         & 30.34  & \textbf{17.19} & 29.37   \\
 10 & 0.15                          & 39.67    & 17.96        & \textbf{30.40}  & 14.01 & 26.66    \\
 10 & 0.20                          & 34.14    & 16.72        & 30.22  & 13.24 & \textbf{30.39 }   \\
\hline
\end{tabular}
\end{small}
\end{adjustbox}
\label{exp:hico_lambda_delta}
\end{table}

\begin{figure*}[ht]
\centering
\includegraphics[width=0.9\linewidth]{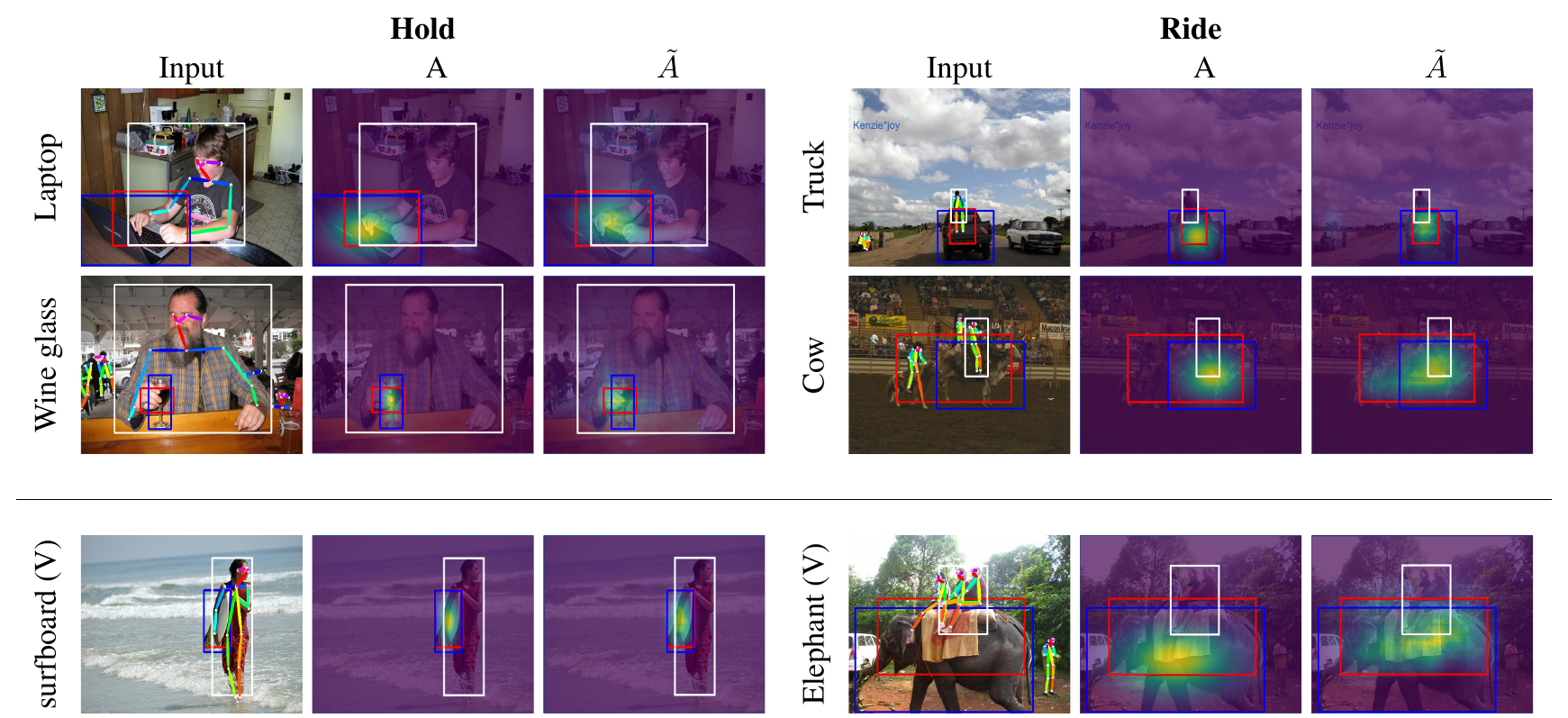}
\caption{(Left) Input image with pose, (middle) ground truth Gaussian attention mask (A) \nj{in yellow}, and (Right) predicted attention map ($\tilde{A}$). 
\textcolor{red}{Red box} is pseudo object box , \textcolor{blue}{blue box} is ground truth and white box indicates the human in \nj{action}. 
(V) \nj{the last row} is the result on the V-COCO dataset. 
Note that a white box is used solely \nj{for visually representing} an acting human in an image and is not used in training on  $D_{T}$.}
\label{exp:example_image}
\end{figure*}

\subsection{Comparison with different $\lambda$ and $\delta$}

In experiments in Table~\ref{exp:hico_lambda_delta}, we focus on verifying the effect of shared parameters such as $\lambda$ in (\ref{eq:total_loss}) and $\delta$ in (\ref{eq:threshold}). 


In top of Table~\ref{exp:hico_lambda_delta}, according to the change of the $\lambda$ in (\ref{eq:total_loss}), the ratio of the loss weight in RRPN is determined. 
When $\lambda$ is zero, due to untrained RRPN, Recall and mAP for $D_{T}(W)$ have the lowest score while mAP for $D_{S}$ has the highest score.
On the other hand, Recall and mAP are the highest at $\lambda = 10$ with performance improvements of 17.45\% and 15.58\% compared to $\lambda = 0$, respectively.
On the contrary, on some levels of $\lambda$, we can see that the performance degradation for not only $D_{T}(W)$ but also $D_{S}$.

This can be understood as an effect of parameter sharing for image feature extractor between RRPN and an object detector.
As mentioned earlier, RRPN is a universal add-on type module which can be adapted to various computer vision tasks. 
To effectively utilize these advantages, we share the backbone network of RRPN and the object detector in consideration of memory efficiency. 
Therefore, the RRPN and the object detector affect each other through the backbone network during training.


In bottom of Table~\ref{exp:hico_lambda_delta}, according to the $\delta$ in (\ref{eq:threshold}), the size of the pseudo bounding box is determined. 
A small $\delta$ makes the size of the boxes increase, while a large $\delta$ makes the box small or disappear. 
As $\delta$ increases, partial information of the object is trained. 
For example, in the case of an apple, only the central part of the apple is trained with high $\delta$, which causes many false positive. 
On the other hand, lowering the threshold of the box, Faster-RCNN is trained not only with an object but also with backgrounds. 
It is interesting that $\delta$ affect differently to both metrics where Recall gets higher when $\delta$ gets smaller but mAP get the highest score when $\delta = 0.1$.
We believe that the RRPN can easily learn objectness with a larger box due to small $\delta$, but classification of objects could be more difficult due to inaccurate localization.
Therefore, too small or too large $\delta$ causes a degradation of mAP, 
and we have found the suitable value, $\delta = 0.1$, through the experiments by selecting the value with the highest performance in $D_{T}(W)$.




\begin{figure}[ht]






  

\centering
\includegraphics[width=\linewidth]{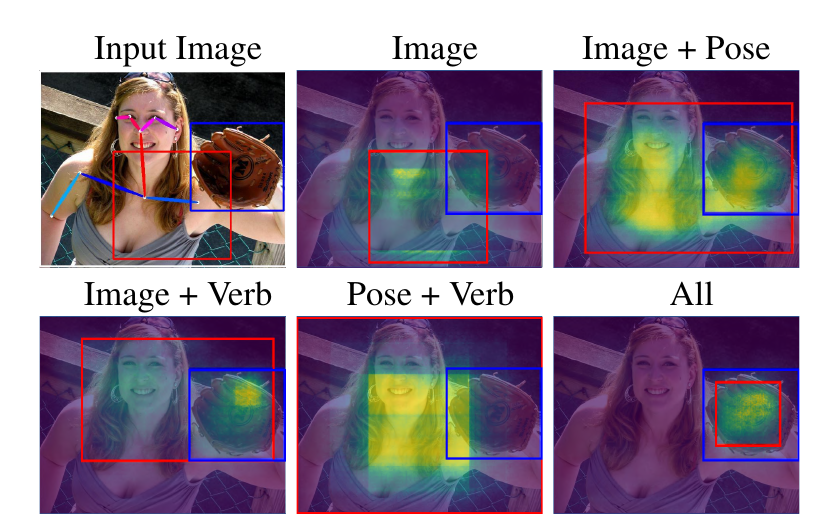}
\vspace{-0.5cm}
\caption{Comparison of predicted attention maps \nj{trained only by the image feature and by various integrated features with [glove, hold].} 
\nj{The} predicted attention maps show different activations depending on the role of each feature map.}
\vspace{-3mm}
\label{exp:img_full}
\end{figure}

\subsection{Qualitative results}
Fig. \ref{exp:example_image} shows the qualitative results of the proposed algorithm on $D_T$.
The first column indicates input images and the last column indicates output attention maps inferred by the corresponding actions.
We can see that RRPN predicts an accurate attention map on unseen object classes in $D_T$.
Furthermore, it can be seen that the pattern of the predicted attention map differs depending on the verb. 
For example, while \textit{`hold'} shows a strong activation value near the human hand, \textit{`ride'} tends to activate at the bottom of a person.
Based on this, we can confirm that the object location can be estimated based on the interaction between the verb and the pose.
The role of the pose can be found in the example of [Truck, Ride].
Despite that two trucks exist in an image, the activation of a truck on which the human is riding shows stronger than the other. 
This can be seen as a contribution of $F_{pose}$ to the object localization.
We also verified the performance of RRPN on the object from a different dataset, V-COCO. 
The RRPN is trained using $D_{S}$ of HICO-DET and predicts the attention map of $D_{T}$ of V-COCO. 
The bottom row shows the predicted attention map on V-COCO. 
We can see that the proposed algorithm can also predict the object location accurately on images even from other datasets.

Fig. \ref{exp:img_full} depicts the comparison of predicted attention map between different feature combinations on [glove, hold].
As described in section 5, the pattern of the resulting attention map can be changed by the combination of features.
Since ``Glove" is an unseen class, backbone has no information to extract reliable feature, 
so that RRPN cannot predict the location of an object accurately using only $F_{img}$.
However, if RRPN is trained using more than two features including $F_{img}$, 
RRPN can infer the location of an object based either on $F_{pose}$ or on $F_{verb}$.
Specifically, $F_{img} + F_{verb}$ predicted a more distinguishable attention map for an object, 
compared to $F_{img} + F_{pose}$ feature map.
Since $F_{img}$ and $F_{pose}$ are extracted from the same image, some of the information can be redundant between two features. 
On the other hand, $F_{verb}$ is able to provide useful information to $F_{int}$ because it is extracted from a different domain, language.
Consequently, the location of an object can be predicted precisely when we use all three features.
On the contrary, if RRPN trained using only $F_{pose}$ and $F_{verb}$ without $F_{img}$, the output attention map only activates around the human. 
Thus, it can be understood that $F_{verb}$ plays a role of providing supplementary information to $F_{img}$ about the object of interest.

\section{Conclusion}
In this paper, we proposed a novel weakly-supervised scheme for object detection problems.
We introduced the RRPN which can universally localize objects in an image with information on human poses and action verbs.
Using transferable knowledge from the RRPN, we can continuously train any object detector for unseen objects with weak verbal supervision describing HOI.
We validated our method based on the results on HICO-DET dataset and the performances show the possibility of our method for a new WSOD training scheme.  
Our work shows sufficient potentials to overcome the inefficiency of the supervised training scheme in recent deep learning.
Also, we can develop our method in the direction to the continual learning since we already suggested a novel method to transfer common knowledge to localize objects with HOI.

\section{Acknowledgments}
\label{sec:Acknowledgments}
This work was supported by Next-Generation Information Computing Development Program through the NRF of Korea (2017M3C4A7077582) and  Promising-Pioneering Researcher Program through Seoul National University(SNU) in 2015.

\bibliographystyle{aaai} 
\bibliography{aaai20}

\newpage
\appendix

\section{Dataset}
We experiment with HICO-DET \cite{chao2018learning} and V-COCO \cite{gupta2015visual} datasets in consideration of HOI.
Since, However, those datasets are not designed for the problem we defined, 
we need to consider some limitations.
Most of them are related to ground truth annotation such as the class or the bounding box and can affect performance evaluation.
We designed the whole experiments in consideration these limitations.
Details of each dataset regarding the limitations are described in following subsections.

\begin{figure}[t]

\centering
\includegraphics[width=0.9\linewidth]{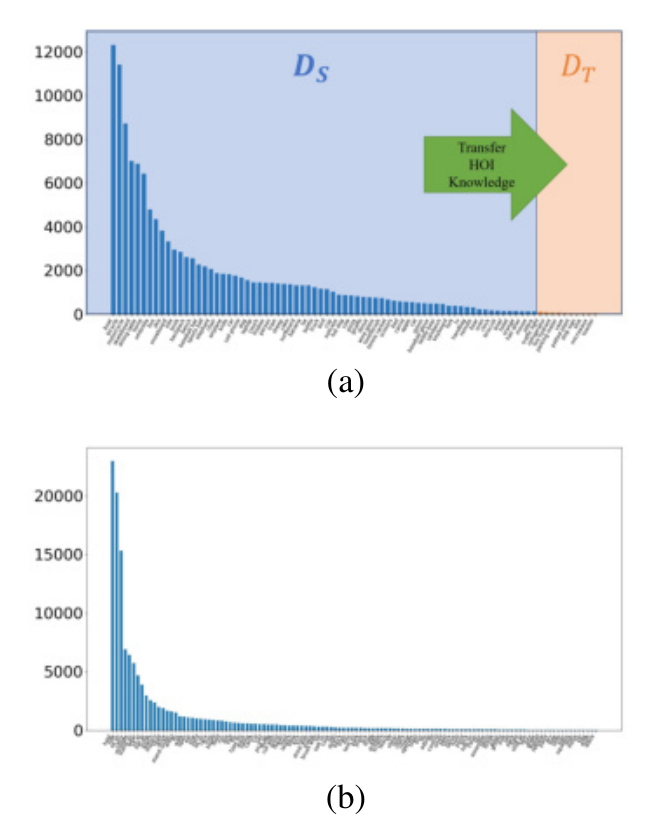}
\caption{The number of tuples according to the objects and actions in HICO-DET. (a) is the number of each object in the tuples and (b) is the number of each action(verb) in the tuples}
\vspace{-2mm}
\label{figure:dataset:hico}
\end{figure}

\begin{table}[ht]
\caption{Comparison the number of tuple between non-rare class(top-10(obj)/5(verb) and rare class(bottom-10(obj)/5(verb)) in HICO-DET}
\label{table:dataset:hico}
\centering
\begin{adjustbox}{max width=\textwidth}
\begin{small}
\begin{tabular}{| c || c | c || c | c |}
\hline
Class                & non-rare class   & Tuple        & rare class & Tuple \\   
\hline
\hline
Obj                  & boat   & 12,311     & traffic light     & 108 \\   
                     & bicycle & 11,422     & refrigerator     & 94 \\   
                     & motorcycle   & 8,719    & fire hydrant     & 82 \\   
                     & skateboard  & 7,015     & parking meter     & 79 \\   
                     & dining table   & 6,880     & vase     & 55 \\   
                     & horse   & 6,435     & potted plant     & 49 \\   
                     & umbrella   & 4,797     &  stop sign   & 47 \\   
                     & bus   & 4,343     &  sink    & 40 \\   
                     & skis   & 3,821     & microwave     & 30 \\   
                     & snowboard   & 3,324     & toaster     & 29 \\   
\hline
Verb                 & hold    & 22,969      & move     & 10 \\  
                     & ride    & 20,289      & zip     & 6 \\  
                     & sit on    & 15,318    & tag     & 6 \\  
                     & carry    & 6,911      & flush     & 6 \\  
                     & straddle    & 6,434   & wave     & 5 \\  
\hline
\end{tabular}
\end{small}
\end{adjustbox}
\end{table}

\subsection{HICO-DET}
\label{sec:hico}
\noindent \textbf{Data Imbalancing}
Table.\ref{table:dataset:hico} and Fig.\ref{figure:dataset:hico} show the frequency of each verb and object for tuples of HICO-DET. 
We can see a large gap between the number of frequently-appearing objects and the number of rarely-appearing objects.
Due to the imbalance of the number of tuples between the object classes, some HOI's, such as [refrigerator, move], are rarely trained on RRPN.
Thus the effectiveness of RRPN on such tuples may not be verified accurately.

For tuples such as [refrigerator, move], RRPN could predict inaccurate $\tilde{O}_{bbox}$, 
so that detection performance on $D_{T}$ could also be degraded.

\gl{For qualitative result, we use 5 verbs and 10 $D_{S}$ and 70 $D_{T}$, other hyperparameters are the same as the basic set up. The reason why we shrink the size of the dataset for the qualitative result is for representing the effectiveness of RRPN more clearly. In other words, due to the inherent limitation of HICO-DET, it is difficult to obtain clear visual results to show the performance of RRPN. We, therefore, exploited non-rare verbs and increased the ratio of $D_{T}$ to see the result for various objects.}

\begin{figure*}[ht]
\centering
\includegraphics[width=\linewidth]{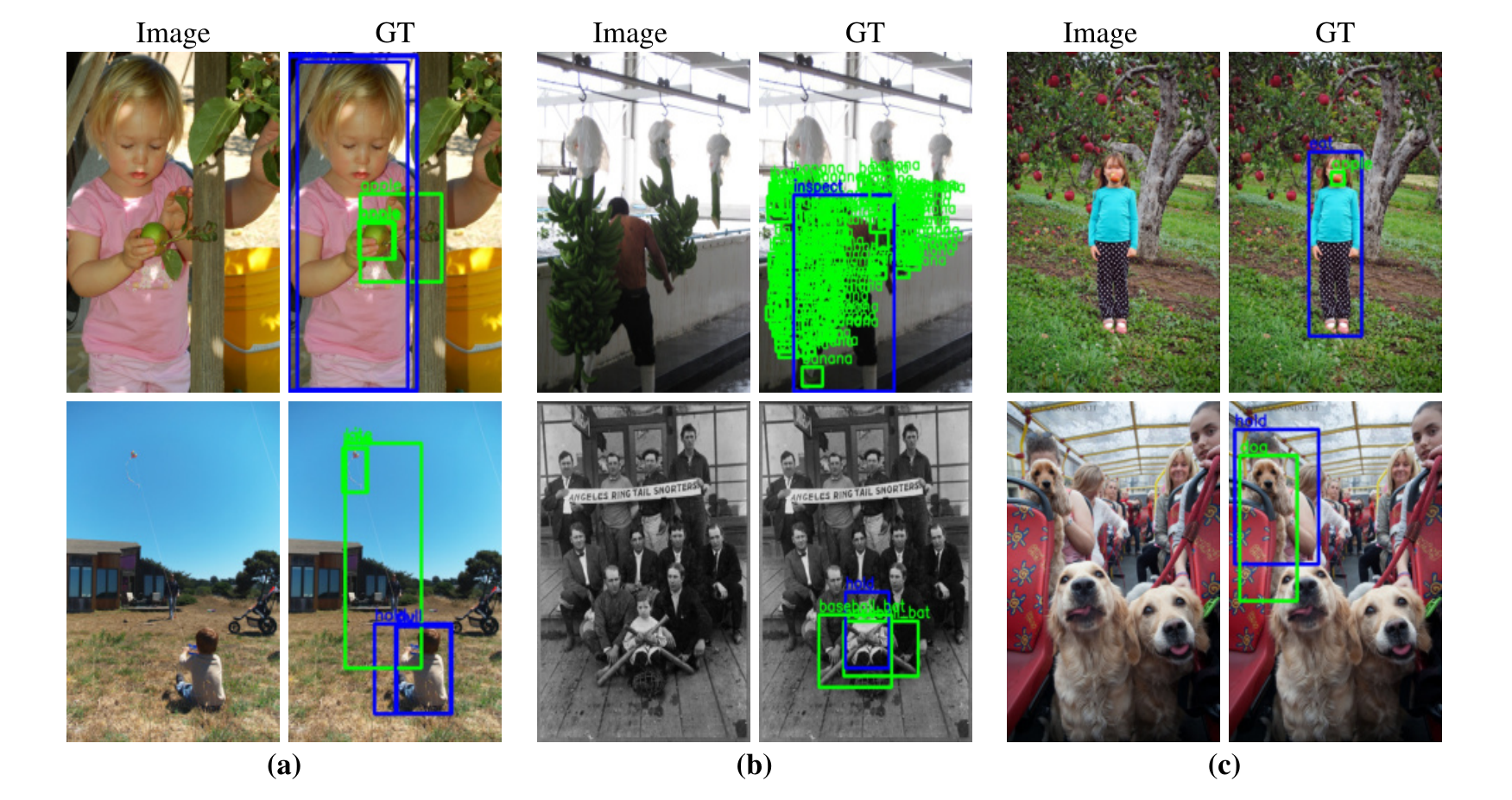}
\caption{Examples of ground truth bounding box annotations of HICO-DET. 
These examples represent (a) bounding boxes are different for the same object (Top : apple, Bottom : kite), (b) overlapped bounding boxes annotation (Top : banana, Bottom : baseball bat) and (c) missing ground truth bounding boxes on an image (Top : apple, Bottom : dog). These insufficient bounding box annotations could induce biased results.}
\label{limitation:hico:gt}
\end{figure*}

\vspace{2mm}
\noindent \textbf{Annotations}
HICO-DET is a dataset for HOI which is expanded for the object detection problem, and the image label is organized as a tuple as shown in Eq. (1).
Thus, multiple ground truth object bounding boxes can be annotated in a single object as shown in Fig.\ref{limitation:hico:gt}.(a).
On the contrary, as depicted in Fig.\ref{limitation:hico:gt}.(b), bounding boxes of multiple objects can also be overlapped each other.
Besides, HICO-DET is not designed only for object detection problem, 
so some object bounding box annotations are missing as shown in Fig.\ref{limitation:hico:gt}.(c).
This is different from the set up of standard object detection problems where only one box is annotated for one object, 
so the result of validation, such as mAP, can be biased.
One way to address the aforementioned problem is integrating multiple object bounding boxes into one.
However, this approach can result in other bias in the results.
This is because, as shown in Fig. \ref{limitation:hico:gt}.(a), the ground truth bounding boxes for a single object are 
not only shown a different pattern depending on verbs but also not identically the same even on the same verb.
Furthermore, if we integrate boxes depending on the IoU scores, 
bonding box labels of some objects can be missing as shown in Fig.\ref{limitation:hico:gt}.(b).

Considering the above-mentioned conditions of HICO-DET, 
experiments in this paper are conducted using the original label of HICO-DET, 
i.e. without intentionally manipulates on the label of HICO-DET.
Obviously, the experimental results can be biased by the effect of multiple ground truth bounding boxes compared to the case of using only one.
Nevertheless, we conducted the whole experiments in the exact same condition and we did not compare our proposed method with other WSOD algorithms.
Therefore we believe that the experimental results presented in the paper 
are reasonable to sufficiently demonstrate the effectiveness of the proposed method.

\subsection{V-COCO}
\label{limitation:vcoco}

V-COCO dataset is made of a subset of the MS-COCO dataset \cite{lin2014microsoft} 
and has 26 actions and 2 object classes (\textit{direct object and instrument}). 
There is a problem of how to separate the source and the target. 
The object classes which are binary have high dependency on verb classes. 
Thus, the verb in the test set does not exist in the training set.
On the other hand, when we transform the two classes into 80 classes, 
there are about 10 empty classes because V-COCO is not designed for object detection. 
In addition, some classes in the test set do not exist in the training set. 
Therefore, we present activation of the tuple with the same verb as HICO-DET.

\section{Implementation details}

The overall structure of the proposed WSOD method consists of RRPN and Faster-RCNN. 
We have used Faster-RCNN with ImageNet pre-trained ResNet-101 model. 
RRPN consists of $f_{img}(I_{i}, \theta_{img})$, $f_{pose}(I_{i}, \theta_{pose})$, and $f_{word}(H_{act}^{j}, \theta_{word})$, 
and each model consists of backbone of Faster-RCNN, OPENPOSE\footnote{https://github.com/CMU-Perceptual-Computing-Lab/openpose} \cite{cao17}, 
and GloVe\footnote{https://github.com/stanfordnlp/GloVe} \cite{Pennington14} as described previously.
We have used the pre-trained OPENPOSE and GloVe models only to extract each feature without further training.

The spatial dimension of the integrated feature map is equal to the output feature map of the Faster-RCNN backbone network, i.e. $40 \times 40$.
$f_{en}$, $f_{de1}$ and $f_{de2}$ consist of 3, 5, 6 convolutional layer, respectively. 
Max pooling and appropriate strides were used to fit the corresponding spatial dimension.
$f_{att}$ is $1 \times 1$ convolution layer followed by a sigmoid layer.
The spatial dimension of the output feature map is the same as the input spatial dimension of the object detector.
Hyper parameters for training RRPN is as follows:
learning rate: 1e-3, optimizer: stochastic gradient descent, weight decay: 1e-4,
momentum: 0.9, batch size: 4, epoch or iteration: 15 epoch (source class), 30 epoch (target class).
Hyper parameters for Faster-RCNN is set as the same as the original paper.

\section{Result}
In this section, we present the performance of our algorithm by further analyzing experimental results for various situations. 

\subsection{The role of features}
We additionally analyze the role of features by conducting the following experiments: 1) $f_{img}$ fix, 2) $F_{pose}$ only, 3) $F_{verb}$ only.
$f_{img}$ fix is the case where we train $D_{T}$ while the parameters of a shared backbone network, which is trained on $D_{S}$, are fixed.
$F_{pose}$ only and $F_{verb}$ only refer the cases where we train RRPN using only $F_{pose}$ and $F_{verb}$, respectively.

Table \ref{exp:supple_feat} represents the results of each experiment. 
Training on $D_{T}$ is conducted using the fixed image feature, $f_{img}$, 
that is already trained on $D_{S}$. So, $f_{img}$ is unable to learn to extract proper image features for $D_{T}$. 
Consequently, mAP decreases not only for the case of weakly supervised learning ($D_T(W)$) but also for the supervised learning approach ($D_T(S)$).

\begin{table}[ht]
\caption{Quantitative result of feature combinations: $f_{img}$ fix, $F_{pose}$ only, and $F_{verb}$ only.
$\ast$ means that the model for the corresponding feature is fixed during the training on $D_{T}$.
(I : Image feature, P : Pose feature, V: Verb feature, $D_{S}$ : Source, $D_{T}$ : Target, W : Weakly supervised object detection, S : Supervised object detection, $\lambda$ = 10, $\delta$ = 0.1)} 
\vspace{-1mm}
\centering
\begin{adjustbox}{max width=\textwidth}
\begin{small}
\renewcommand{\tabcolsep}{4pt}

\begin{tabular}{| c | c | c || c | c || c | c | c |}
\hline
I & P & V & \multicolumn{2}{c||}{\footnotesize{Recall@.5} \scriptsize{(RRPN)}}       & \multicolumn{3}{c|}{mAP@.5 \footnotesize{(Faster-RCNN)}} \\
\cline{4-8}                    & &     & $D_{S}$ ($\%$) & $D_{T}$ ($\%$) & $D_{S}$         & $D_{T}$(W)  & $D_{T}$(S)\\
\hline
\hline
\checkmark$^{\ast}$ & \checkmark & \checkmark    & 47.69    & 28.64        & 30.34  & 10.58  &22.90   \\
\hline
& \checkmark &                          & 22.05    & 17.34        & 32.14  & 5.24 & 28.20   \\
&  & \checkmark                         & 16.78    & 10.06        & 31.24  & 3.54 & 27.20   \\
& \checkmark & \checkmark               & 20.28    & 11.00        & 31.60  & 4.33 & 28.50   \\
\hline
\end{tabular}
\end{small}
\end{adjustbox}
\label{exp:supple_feat}
\end{table}

The experiment results on $F_{pose}$ only and $F_{verb}$ only, regarding $D_{T}$(W) are 
significantly lower than other feature combinations that use $F_{img}$. 
It implies that using a single modality of $F_{pose}$ or $F_{verb}$ only could be insufficient to predict an accurate object location.
In other words, to accurately predict an object location from HOI, 
one needs to provide not only human pose and corresponding action (verb) but also information on the object of interest contained in the image feature.
Meanwhile, since $F_{verb}$ is copied to every spatial positions to have the same value regardless of positions, 
the experiment results using only $F_{verb}$ have the lowest performance compared to other feature combinations.
It is strange that the performance decreases when both $F_{pose}$ and $F_{verb}$ are used simultaneously than when only $F_{pose}$ is used.

\subsection{An effect of a ratio between $D_{S}$ and $D_{T}$}

We experiment to verify the performance according to the different ratio of the source classes and the target classes. 
We experiment with reducing 10 source classes repeatedly and the number of the verbs is fixed to 116 excluding 'no interaction'.

Our result of the basic ratio in the top of Table~\ref{exp:data_bal} shows the highest performance in both Recall and mAP in the target test. 
On the contrary, the result of the lowest ratio of the source classes in the bottom of Table~\ref{exp:data_bal} shows the highest performance in both Recall and mAP in the source test. 
As the ratio of the source classes decreases, the task at the source is relatively easier and improves performance because of the reduction of classes. 
On the other hand, in the target, the task becomes relatively difficult and decreases the performance because of the increase of classes.
Other reason can be attributed to the reduced number of training examples as the number of source classes decreases.

\begin{table}[ht]
\caption{Comparison of quantitative result according to the different ratio of the source classes and the target classes 
($D_{S}$ : Source, $D_{T}$ : Target, W : Weakly supervised object detection, S : Supervised object detection, $\lambda$ = 10, $\delta$ = 0.1, the number of actions (verb) = 116)}
\label{exp:data_bal}
\centering
\begin{adjustbox}{max width=\textwidth}
\renewcommand{\tabcolsep}{5pt}
\begin{small}

\begin{tabular}{| c | c || c | c || c | c | c |}
\hline
\multicolumn{2}{|c||}{\# of classes}    & \multicolumn{2}{c||}{\footnotesize{Recall@.5} \scriptsize{(RRPN)}}      & \multicolumn{3}{c|}{mAP@.5 \footnotesize{(Faster-RCNN)}}\\   
\hline
$D_{S}$ & $D_{T}$          &  $D_{S}$ ($\%$) & $D_{T}$ ($\%$) & $D_{S}$         & $D_{T}$(W)  & $D_{T}$(S)\\   
\hline
\hline
70 & 10             & 47.69      & \textbf{28.64}       & 30.34     & \textbf{17.19}      & \textbf{29.37}\\
\hline
60 & 20             & 47.26      & 25.35       & 31.51     & 11.52      & 27.15\\
50 & 30             & 46.47      & 27.34       & 33.69     & 12.76      & 24.36\\
40 & 40             & 45.73      & 23.45       & 34.55     & 9.63       & 25.16\\
30 & 50             & \textbf{48.53}      & 20.14       & \textbf{40.29}     & 9.46       & 21.97\\
\hline
\end{tabular}
\end{small}
\end{adjustbox}
\end{table}

\subsection{More results}
Note that experiments for the qualitative results in the main paper and supplement material
are conducted on a smaller dataset compared to the basic set up for the quantitative results. 
For qualitative result, we use 5 verbs and 10 $D_{S}$ and 70 $D_{T}$, other hyperparameters are the same as the basic set up.
The reason why we shrink the size of the dataset for the qualitative result is for representing the effectiveness of RRPN more clearly.
In other words, due to the inherent limitation of HICO-DET (see section \ref{sec:hico}), 
it is difficult to obtain clear visual results to show the performance of RRPN. 
We, therefore, exploited non-rare verbs and increased the ratio of $D_{T}$ to see the result for various objects.
The quantitative results on the smaller dataset are represented in Table \ref{supple:small1} and \ref{supple:small2}.
Obviously, the performance of $D_T(W)$ decreases. 
Since, however, the quantitative results on the smaller dataset show analogous tendency to the results on the basic set up, 
we believe the qualitative results on the smaller dataset can be used to represent the performance of RRPN.

\begin{table}[ht]
\caption{Comparison of quantitative result of different feature combination. (I : Image feature, P : Pose feature, V: Verb feature, $D_{S}$ : Source, $D_{T}$ : Target, W : Weakly supervised object detection, S : Supervised object detection, $D_{S}$ = 10, $D_{T}$ = 70, $\lambda$ = 10, $\delta$ = 0.1)}
\vspace{-1mm}
\centering
\begin{adjustbox}{max width=\textwidth}
\begin{small}
\renewcommand{\tabcolsep}{4pt}

\begin{tabular}{| c | c | c || c | c || c | c | c |}
\hline
I & P & V & \multicolumn{2}{c||}{\footnotesize{Recall@.5} \scriptsize{(RRPN)}}       & \multicolumn{3}{c|}{mAP@.5 \footnotesize{(Faster-RCNN)}} \\
\cline{4-8}                    & &     & $D_{S}$ ($\%$) & $D_{T}$ ($\%$) & $D_{S}$         & $D_{T}$(W)  & $D_{T}$(S)\\
\hline
\hline
\checkmark & \checkmark & \checkmark    & 48.62    & \textbf{19.99}        & 46.38  & \textbf{10.06} & 27.17   \\
\hline
\checkmark &  &                         & 47.69    & 17.66        & 45.75  &  7.23 & 26.33   \\
 & \checkmark &                         & 29.54    & 16.23        & 47.18  &  4.72 & \textbf{28.03}   \\
 &  & \checkmark                        & 22.72    & 11.31        & \textbf{47.73}  &  2.48 & 27.03   \\
\hline
\checkmark & \checkmark &               & 48.34    & 16.84        & 46.54  &  7.07 & 25.42   \\
\checkmark &  & \checkmark              & \textbf{48.91}    & 18.25        & 47.19  &  8.12 & 25.74   \\
 & \checkmark & \checkmark              & 12.66    & 13.16        & 47.69  &  2.51 & 26.14   \\
\hline
\end{tabular}
\end{small}
\end{adjustbox}
\label{supple:small1}
\end{table}

\begin{figure*}[ht]
\centering
\includegraphics[width=\linewidth]{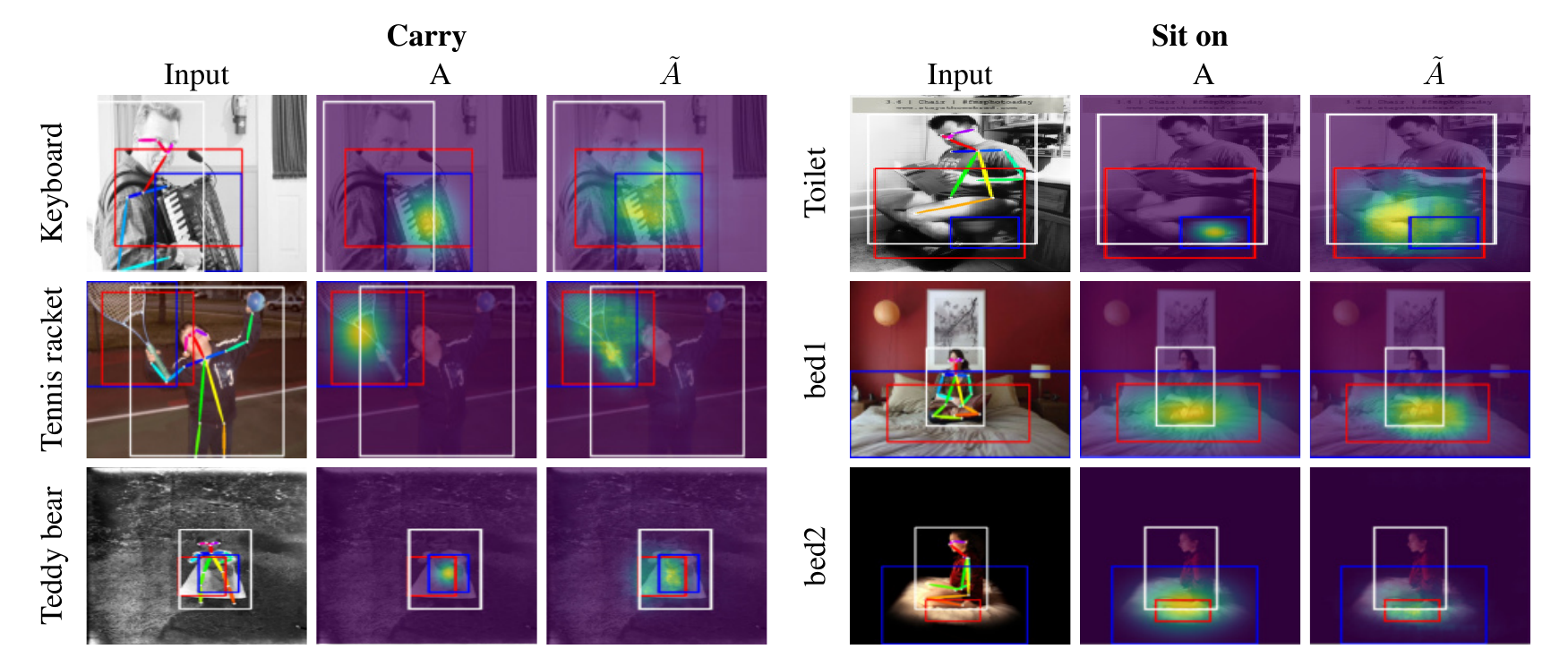}
\caption{The qualitative results on \textit{carry} and \textit{sit on} about three different objects each
Qua(Left) Input image with pose, (middle) ground truth Gaussian attention mask, and (Right) predicted attention map.
\textcolor{red}{red box} is pseudo object box , \textcolor{blue}{blue box} is ground truth and white box indicates the human in acting.
Note that a white box is used to solely on represent an acting human in an image and is not used in training
on the target class $D_{T}$}
\label{supple:example_image}
\end{figure*}

Fig. \ref{supple:example_image} shows the qualitative results on $D_{T}$ especially regarding \textit{carry} and \textit{sit on}.
The pattern of the predicted attention maps differs depending on the verb: 
while \textit{carry} focuses on near the human hand, \textit{sit on} activates at the bottom of a human.
These results are analogous to results of \textit{hold} and \textit{ride} in the main paper.

\begin{figure*}[ht]
\centering
\includegraphics[width=\linewidth]{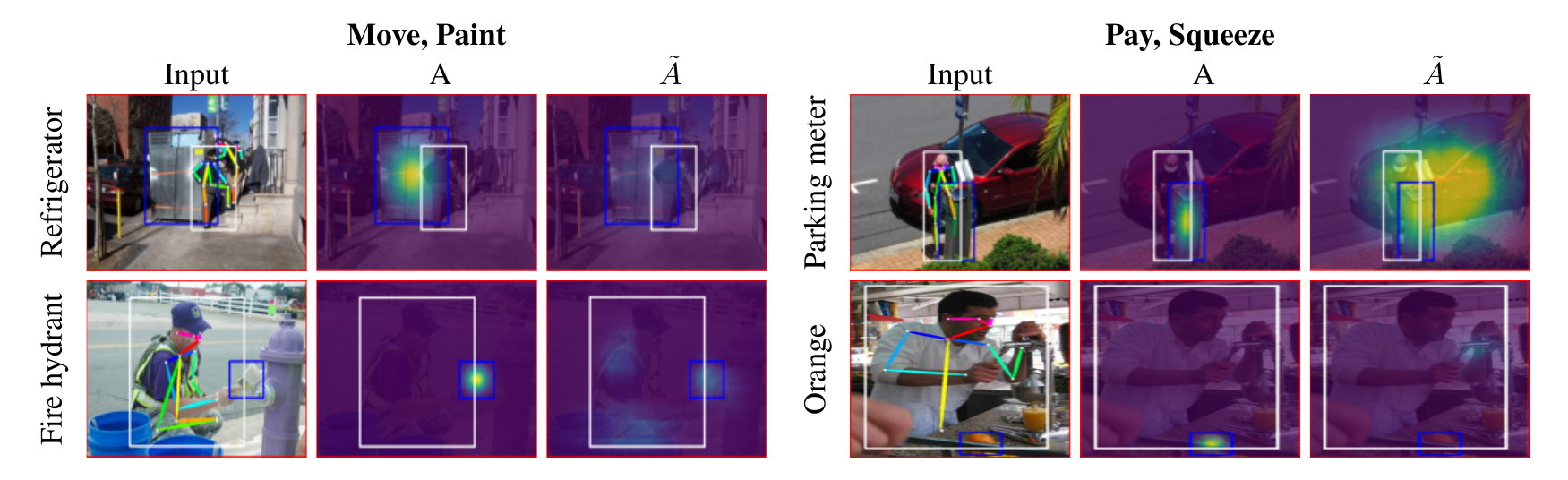}
\caption{Unsuccessful results. Box color notation is the same as Fig. \ref{supple:example_image}.
Incorrect attention maps are predicted due to the influence of rarely-appearing tuple which is insufficient trained.}
\label{supple:good}
\end{figure*}

On the contrary, the predicted attention maps on rarely appearing verbs are inaccurate as shown in Fig. \ref{supple:good}.
This is because HOI related to rarely appearing verbs could not be sufficiently trained.

\begin{figure*}[ht]




\centering
\includegraphics[width=0.7\linewidth]{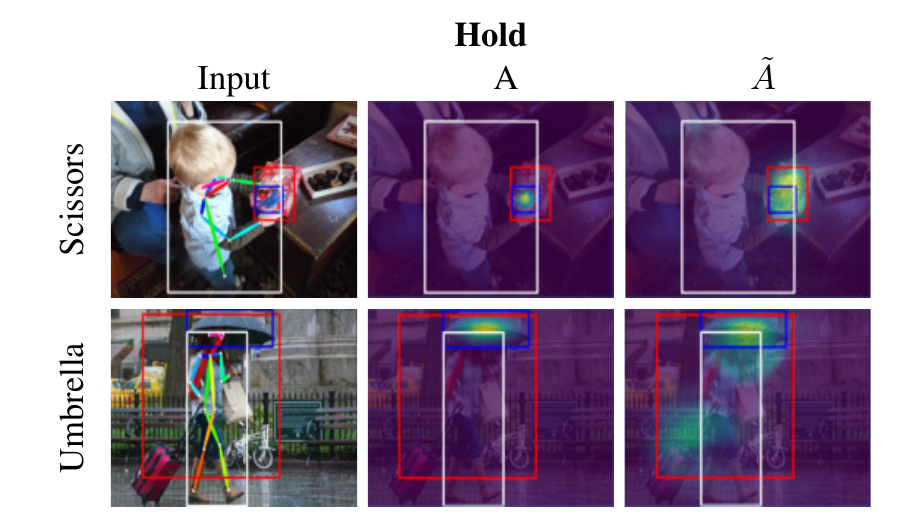}
\caption{Unsuccessful results. Box color notation is the same as Fig. \ref{supple:example_image}. 
Inaccurate object localization occurs when 
(top) two objects are overlapping (bottom) or the human doing the same action for multiple objects.}
\label{exp:bad_case}
\end{figure*}

Fig. \ref{exp:bad_case} also shows some examples of unsuccessful results.
The verb and object class labels are annotated as [hold, scissors] and [hold, umbrella], respectively.
The predicted attention maps are focused on the box instead of the scissors, 
and on both umbrella and suitcase rather than only the umbrella. 
Since RRPN predicts the location of an object according to human and its corresponding action, 
inaccurate attention maps can be generated when multiple objects are involved in the same action.

\begin{table}[ht]
\caption{Comparison of quantitative result of different attention loss balance and box threshold ($D_{S}$ : Source(10), $D_{T}$ : Target(70), W : Weakly supervised object detection, S : Supervised object detection)}
\vspace{-1mm}
\centering
\begin{adjustbox}{max width=\textwidth}
\renewcommand{\tabcolsep}{5pt}
\begin{small}
\begin{tabular}{|c | c || c | c || c | c | c |}
\hline
\multicolumn{2}{|c||}{parameter}  &\multicolumn{2}{c||}{\footnotesize{Recall@.5} \scriptsize{(RRPN)}}       & \multicolumn{3}{c|}{mAP@.5 \footnotesize{(Faster-RCNN)}} \\
\hline
$\lambda$ & $\delta$    & $D_{S}$ ($\%$) & $D_{T}$ ($\%$)  & $D_{S}$         & $D_{T}$(W)  & $D_{T}$(S) \\
\hline
\hline
0 & 0.1                             & 12.66    & 13.16        & \textbf{47.71}  &  2.53 & 27.62   \\
\hline
 1 & 0.1                            & 47.85    & 17.07        & 45.00  &  6.80 & \textbf{27.73}   \\
 5 & 0.1                            & 47.96    & 18.82        & 46.15  &  7.88 & 27.25   \\
 10 & 0.1                           & \textbf{48.62}    & \textbf{19.99}        & 46.38  & \textbf{10.06} & 27.17   \\
 15 & 0.1                           & 47.52    & 17.94        & 44.12  &  8.95 & 26.99   \\
 20 & 0.1                           & 44.31    & 15.30        & 43.03  &  7.52 & 27.01   \\
\hline
\hline
 10 & 0.05                          & \textbf{50.76}    & \textbf{21.83}        & \textbf{46.98}  &  9.14 & \textbf{27.27}   \\
 10 & 0.10                          & 48.62    & 19.99        & 46.38  & \textbf{10.06} & 27.17   \\
 10 & 0.15                          & 42.76    & 11.98        & 46.03  &  8.50 & 26.68   \\
 10 & 0.20                          & 23.68    &  7.06        & 43.94  &  6.09 & 27.71   \\
 10 & 0.30                          &  8.11    &  4.49        & 45.82  &  2.85 & 26.69   \\
\hline
\end{tabular}
\end{small}
\end{adjustbox}
\label{supple:small2}
\end{table}

\newpage
\newpage

\end{document}